\documentclass[conference]{IEEEtran}
\IEEEoverridecommandlockouts
\usepackage{cite}
\usepackage{amsmath,amssymb,amsfonts}
\usepackage{algorithmic}
\usepackage{graphicx}
\usepackage{textcomp}
\usepackage{xcolor}
\usepackage{multirow}
\usepackage{multicol}
\usepackage{tikz}

\usepackage{caption}
\usepackage{subcaption}
\usepackage{booktabs}
\usepackage{hyperref}

\usepackage{pifont}
\newcommand{\xmark}{\ding{56}}%
\def\BibTeX{{\rm B\kern-.05em{\sc i\kern-.025em b}\kern-.08em
    T\kern-.1667em\lower.7ex\hbox{E}\kern-.125emX}}
\begin{document}
\captionsetup[table]{skip=0pt}
\captionsetup[figure]{skip=0pt}
\title{XPert: \underline{Per}ipheral Circui\underline{t} \& Neural Architecture Co-search for Area and Energy-efficient \underline{X}bar-based Computing 

}

\author{Abhishek Moitra, Abhiroop Bhattacharjee, Youngeun Kim and Priyadarshini Panda\\
Yale University, New Haven, CT, 06511, USA} \vspace{-10mm}

\maketitle 

\begin{abstract}
The hardware-efficiency and accuracy of Deep Neural Networks (DNNs) implemented on In-memory Computing (IMC) architectures primarily depend on the DNN architecture and the peripheral circuit parameters. It is therefore essential to holistically co-search the network and peripheral parameters to achieve optimal performance. To this end, we propose XPert, which co-searches network architecture in tandem with peripheral parameters such as the type and precision of analog-to-digital converters, crossbar column sharing and the layer-specific input precision using an optimization-based design space exploration. Compared to VGG16 baselines, XPert achieves 10.24$\times$ (4.7$\times$) lower EDAP, 1.72$\times$ (1.62$\times$) higher TOPS/W, 1.93$\times$ (3$\times$) higher TOPS/mm$^2$ at 92.46\% (56.7\%) accuracy for CIFAR10 (TinyImagenet) datasets. The code for this paper is available at \textcolor{blue}{https://github.com/Intelligent-Computing-Lab-Yale/XPert.}
\end{abstract}

\vspace{-3mm}

\section{Introduction}
Deep Neural Networks (DNNs) have achieved state-of-the-art performance in large-scale image recognition tasks \cite{alzubaidi2021review, guo2016deep}. In the context of energy-efficient deep learning, In-Memory Computing (IMC) architectures are a promising alternative for overcoming the memory wall bottleneck of traditional von-Neumann computing \cite{yu2021compute, sebastian2020memory}. However, in the recent development of IMC architectures, it is found that peripheral circuits are the significant contributors to the energy and area overhead \cite{yuan2021tinyadc, saxena2022towards, jiang2021analog, qiu2018peripheral}. 

To this end, several hardware-algorithm co-design approaches have employed algorithm-level optimization techniques to reduce the peripheral circuit overheads. In \cite{yuan2021tinyadc, huang2021mixed}, the authors have employed pruning and quantization to reduce the analog-to-digital-converter (ADC) precision and thereby, decrease the ADC area and energy overhead. In \cite{saxena2022towards}, the authors employ binary partial-sum quantization-aware training to realise sense amplifier-based SRAM IMC implementations. Other co-design based works \cite{jiang2021analog, qiu2018peripheral} have leveraged hardware-level optimization in IMC architectures to reduce the peripheral circuit overhead. The authors in \cite{jiang2021analog} propose compact ADC architectures to achieve low area and high throughput implementations. In \cite{qiu2018peripheral}, the authors reuse crossbar peripheral circuits over multiple convolution layers to maximize energy efficiency. Although the above mentioned works achieve good hardware efficiency, they have optimized the DNN or hardware in isolation without considering the co-dependence between the DNN architecture and the IMC hardware. Note, we use the terminology DNN architecture (\textit{DNNArch}) to denote different layer configurations of a DNN such as channel depth, kernel size, layer-depth among others. 

To address this, a line of works such as \cite{jiang2020device, negi2022nax} have employed IMC-aware \textit{DNNArch} search. To achieve high energy efficiency, the authors in \cite{jiang2020device} co-searched \textit{DNNArch} along with layer-specific quantization without considering any hardware-specific design parameters (for example, ADC precision, crossbar-size among others). A recent work NAX \cite{negi2022nax} proposed co-searching \textit{DNNArch} with layer-specific crossbar sizes to maximize the energy-efficiency and DNN accuracy. However, the above works have the following drawbacks: 1) They explore a very small hardware design space (e.g.  weight/activation quantization in \cite{jiang2020device} or crossbar size in \cite{negi2022nax}) during the co-search which might result in sub-optimal solutions. 2) They optimize for energy and accuracy without considerations for area overhead.

\begin{table}[t]
    \centering
    \captionsetup{justification=centering}
    \caption{Table showing empirical trends in energy, delay, area and accuracy of IMC-implemented DNNs. \xmark-- metric is not affected. }
    \label{tab:design_parameter}
    \begin{tabular}{|c|c|c|c|c|} \hline
        \textbf{Design Parameter} & \textbf{Energy} & \textbf{Delay} & \textbf{Area} & \textbf{Accuracy} \\ \hline
        Channel Depth (CD) $\uparrow$ & $\uparrow$ & $\uparrow$ & $\uparrow$ & $\uparrow$ \\ \hline
        ADC Precision (AP) $\uparrow$ & $\uparrow$ & $\uparrow$ & $\uparrow$ & $\uparrow$\\ \hline
        Input Precision (IP) $\uparrow$ & $\uparrow$ & $\uparrow$ & \xmark & $\uparrow$ \\ \hline
        Column Sharing (CS) $\uparrow$& $\uparrow$ & $\uparrow$ & $\downarrow$ & \xmark \\ \hline 
    \end{tabular}
    
\end{table}
\begin{figure}[t]
    \centering
    \includegraphics[width=\linewidth]{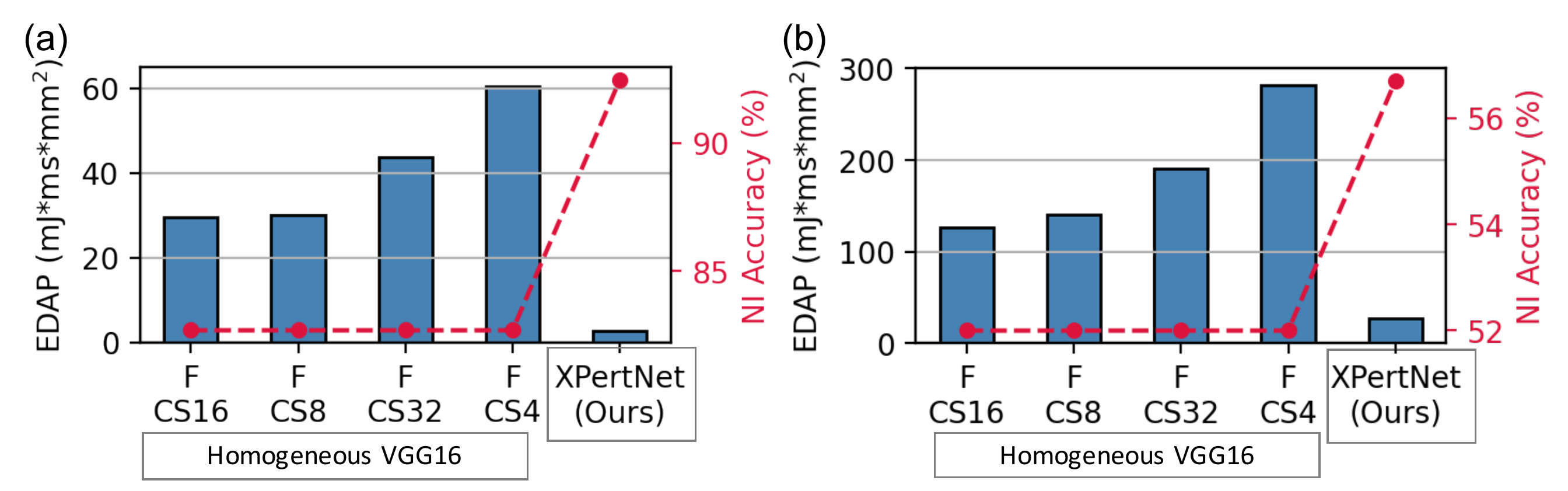}
    \caption{EDAP and non-ideal (NI) accuracy comparison of our XPertNet (having layer-specific peripheral circuit configurations) and VGG16 networks (having homogeneous peripheral circuit configuration across all layers) for (a) CIFAR10 and (b) TinyImagenet datasets. F- Flash ADC, CS8- 8 crossbar columns shared per ADC. All networks are implemented on 64x64 4-bit RRAM crossbars. VGG16 networks are implemented with AP=6 and IP=8.}
    \label{fig:first_result_fig}
\end{figure}

Table \ref{tab:design_parameter} shows different \textit{DNNArch} (layer-specific channel depths (CD)) and crossbar peripheral circuit (ADC Precision (AP), Input Precision (IP) and Column Sharing (CS)) parameters that majorly contribute to the energy, delay, area and accuracy of any IMC-implemented DNN. We define column sharing as the number of crossbar columns shared by one ADC (more details in Section \ref{sec:background_motivation}). In addition to the design parameters shown in Table \ref{tab:design_parameter}, the ADC type (can be either Successive Approximation Register (SAR) or Flash type) plays an important role towards the hardware efficiency and hence is also considered as a design parameter (see Section \ref{sec:background_motivation}). The choice of each design parameter entails different cost and benefits in terms of hardware-efficiency and DNN accuracy. For example as seen in Table \ref{tab:design_parameter}, higher CD, AP and IP values ($\uparrow$) incur higher DNN accuracy ($\uparrow$) while increasing the energy/delay and area consumption ($\uparrow$). A higher CS value decreases area and increases energy/delay. Similarly, among SAR and Flash type ADCs, Flash ADCs are more energy-efficient while being area-inefficient compared to SAR ADCs. 
Given the patent co-dependence between different design parameters, it is essential to perform a holistic co-search in the \textit{DNNArch} and peripheral circuit design space to achieve an optimal solution. 

To this end, we propose XPert, a fast network architecture and crossbar peripheral circuit co-search framework for area and energy-efficient IMC-implementations. XPert uses a dual-phase co-search approach wherein Phase1 is catered towards searching a \textit{DNNmodel} with optimal layer-specific CD, CS and ADC Type (AT) parameters to achieve the lowest delay under a given area constraint. To further optimize the energy-efficiency and accuracy, in Phase2, XPert searches for layer-specific AP and IP configurations on the Phase1-searched \textit{DNNmodel}. Note, we use the terminology \textit{DNNmodel} to denote a DNN with layer-specific architecture and hardware configuration namely CD, CS, AT, AP and IP. Fig. \ref{fig:first_result_fig} shows that XPertNet (the optimal \textit{DNNmodel} obtained from Phase1 and Phase2 co-search) achieves the lowest energy-delay-area product (EDAP) compared to different homogeneous VGG16 implementations. In summary, we make the following contributions:

\begin{enumerate}

    \item We develop the XPert framework to perform fast \textit{DNNArch}-peripheral circuit design space co-search to achieve energy and area-efficient DNN implementations with high classification accuracy. During the co-search, XPert uses XPertSim-Pytorch for real-time area and delay assisted \textit{DNNmodel} search. XPerSim-Pytorch uses a backend XPertSim-C++: a cycle-accurate hardware evaluation platform to support IMC-implemented DNNs with heterogeneous peripheral circuit configurations. 
    
    \item We perform extensive evaluations on CIFAR10 and TinyImagenet datasets and observe that XPertNet (with 50mm$^2$ area) achieves 10.2$\times$ and 4.76$\times$ lower EDAP compared to the most energy-efficient homogeneous VGG16 network (F+CS16+IP8+AP6 in Fig. \ref{fig:first_result_fig}) for CIFAR10 and TinyImagenet, respectively. XPertNet also achieves 9.8\% (CIFAR10) and 4.7\% (TinyImagenet) higher NI accuracy (includes quantization and device variation noise due to analog crossbar computing) compared to baseline VGG16 implementations.
    

    \item Furthermore, we show that XPert can attain optimal accuracy and EDAP under extremely small area constraints. At 20mm$^2$, XPertNet achieves an EDAP (accuracy) of 4.96 mJ*ms*mm$^2$ (92.1\%) and 40.4 mJ*ms*mm$^2$ (56.2\%) on CIFAR10 and TinyImagenet datasets, respectively. 
\end{enumerate}

\section{Background and Motivation}
\label{sec:background_motivation}
\begin{figure}[t]
    \centering
    \includegraphics[width=1\linewidth]{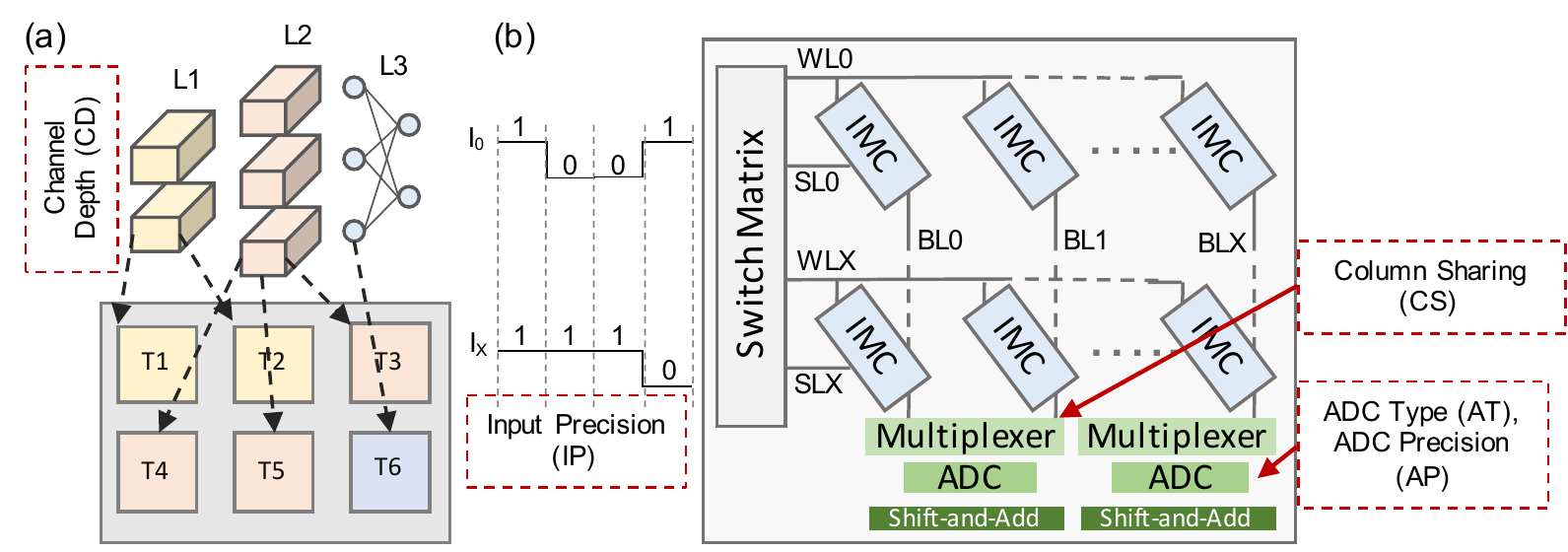}
    \caption{Figure showing (a) the DNN mapping on a tiled-IMC architecture. Each Tile (T1-T6) contains several processing engines (PE) with each PE containing multiple IMC crossbars \cite{peng2019dnn+} (b) the IMC crossbar with different peripheral circuit parameters. The \textit{DNNArch} and peripheral circuit parameters enclosed in red boxes are co-searched in XPert.}
    \label{fig:xbar_background}
\end{figure}
\begin{figure}[t]
    \centering
    \includegraphics[width=0.7\linewidth]{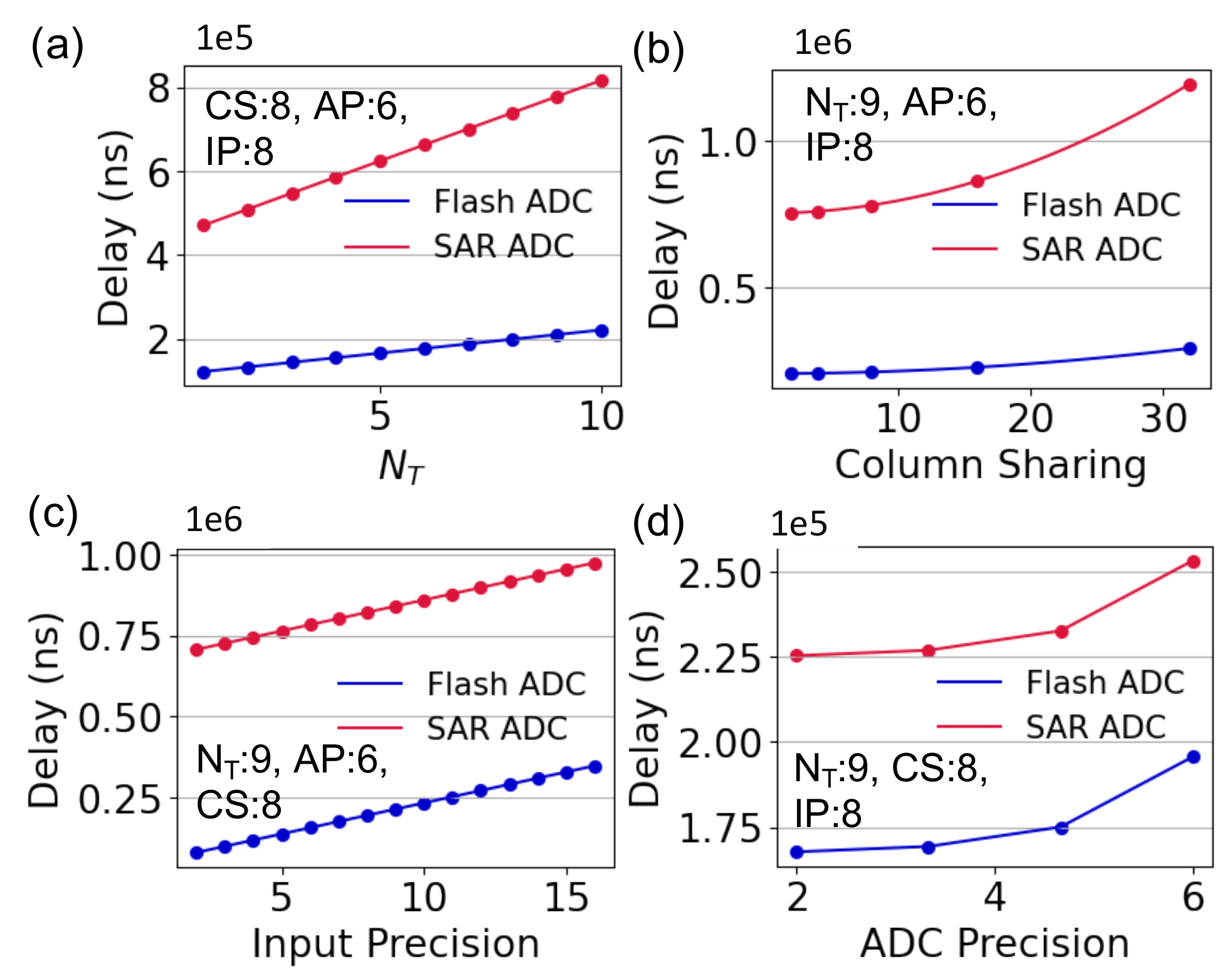}
    \caption{Figure showing the variation of delay with different (a) $N_T$ at fixed CS=8, AP=6 and IP=8 (shown inside the figure) (b) CS (c) IP and (d) AP values. The implementations are based on 64$\times$64 4-bit RRAM crossbars. Note, the energy follows similar trends and not shown for brevity.}
    \label{fig:latency_vs_params}
\end{figure}

\begin{figure}[t]
    \centering
    \includegraphics[width=1\linewidth]{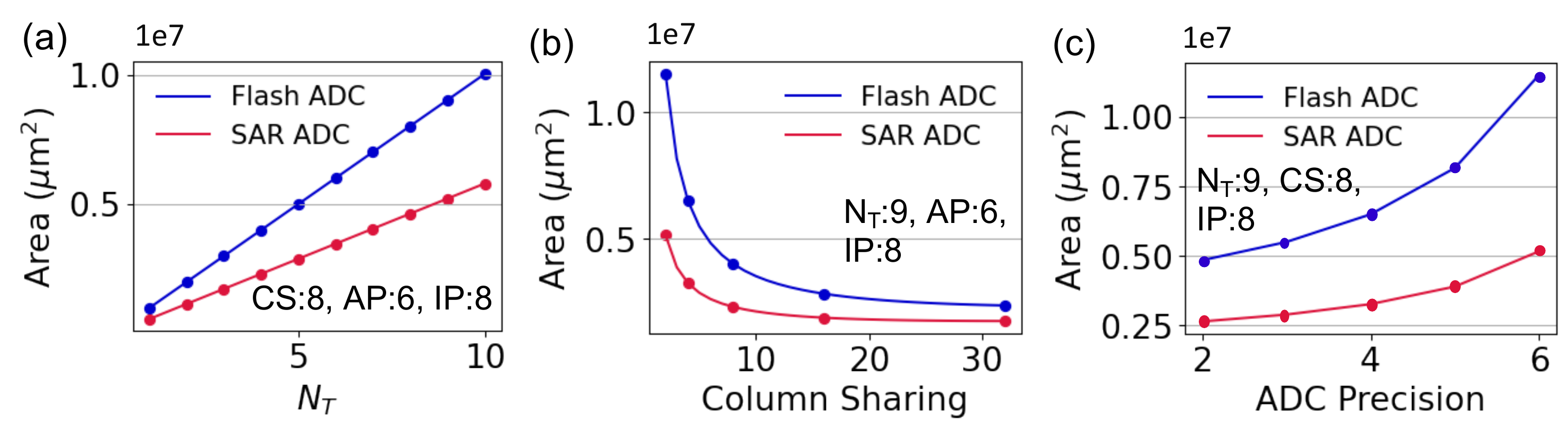}
    \caption{Figure showing the variation of area with different (a) $N_T$ at fixed CS=8, AP=6 and IP=8 (shown inside figure) (b) CS and (c) AP values. The implementations are based on 64$\times$64 4-bit RRAM crossbars.}
    \label{fig:area_vs_params}
\end{figure}
The energy, delay, area and accuracy of IMC-implemented DNNs majorly depend on the following \textit{DNNArch} and peripheral circuit parameters. 
\begin{enumerate}
    \item \textbf{Channel Depth (CD):} In a standard IMC architecture \cite{peng2019dnn+, shafiee2016isaac}, individual layers are mapped on multiple tiles as shown in Fig. \ref{fig:xbar_background}a. The number of tiles ($N_{T,l}$) required for any layer $l$ is majorly dictated by the layer's input ($CD_{in,l}$) and output ($CD_{out,l}$) channel depths. $N_{T,l}$ also depends on factors like the kernel size ($k_l$), crossbar size ($X$) and the number of crossbars (\#Xbars) per tile as shown in Eq. \ref{eq:tile_mapping}. 
    \begin{equation}
    N_{T,l} = \frac{ceil(\frac{CD_{in, l}\times k_l^2}{X}) \times ceil(\frac{CD_{out,l}}{X})}{\#Xbars / Tile}.
    \label{eq:tile_mapping}
\end{equation}
To capture representative features during training and achieve high accuracy, higher CD is required \cite{cai2018proxylessnas}. However, higher CD entail higher $N_{T}$ causing higher delay and area as seen in Fig. \ref{fig:latency_vs_params}a and Fig. \ref{fig:area_vs_params}a, respectively.

\item \textbf{Column Sharing (CS): } As seen in Fig. \ref{fig:xbar_background}b multiplexers at the end of each crossbar facilitate sharing of ADCs and Shift-and-Add circuits among multiple crossbar columns. As shown in Fig. \ref{fig:area_vs_params}b, higher CS requires less ADCs per crossbar and therefore reduces the area. However, higher CS also increases the crossbar read cycles \cite{peng2019dnn+} resulting in higher delay as in Fig. \ref{fig:latency_vs_params}b. 

\item \textbf{Input Precision (IP):} To reduce the overhead of digital-to-analog converters (DAC), standard IMC architectures \cite{peng2019dnn+} employ input bit-serialization wherein a multi-bit input is encoded into a bit-stream processed over multiple crossbar read cycles. For example, a 4-bit input is processed over 4 cycles as seen in Fig. \ref{fig:xbar_background}b. Higher input precision leads higher accuracy and higher delay due to higher crossbar read cycles as seen in Fig. \ref{fig:latency_vs_params}c. 

\item \textbf{ADC Type (AT) and Precision (AP):} In IMC architectures, two types of ADC are predominantly used: Flash type and SAR type. A k-bit Flash-type ADC is composed of $2^{k}-1$ cascaded comparators. Flash ADCs entail extremely low delay at the cost of higher area overhead. A Successive Approximation Register (SAR) ADC employs a single comparator and a DAC to perform bit-by-bit comparison over multiple clock cycles. Due to a single comparator, the area overhead of SAR ADC is significantly less compared to Flash ADC while the delay is much higher as seen in Fig. \ref{fig:latency_vs_params} and Fig. \ref{fig:area_vs_params}. Higher AP is essential for higher accuracy of IMC-implemented DNNs. However, higher AP entails higher cascading and clock cycles/DAC precision in case of Flash and SAR ADCs, respectively. This increases both the delay and area as shown in Fig. \ref{fig:latency_vs_params}d and Fig. \ref{fig:area_vs_params}c.
\end{enumerate}
Evidently, the energy, delay, area and accuracy of IMC-implemented DNNs are functions of CD, CS, AT, AP and IP. Therefore, to achieve an optimal solution, there is a need for holistic \textit{DNNArch} and peripheral circuit co-search. 

\section{Methodology}
\begin{figure*}[t]
    \centering
    \begin{tabular}{c}
         \includegraphics[width=0.9\textwidth]{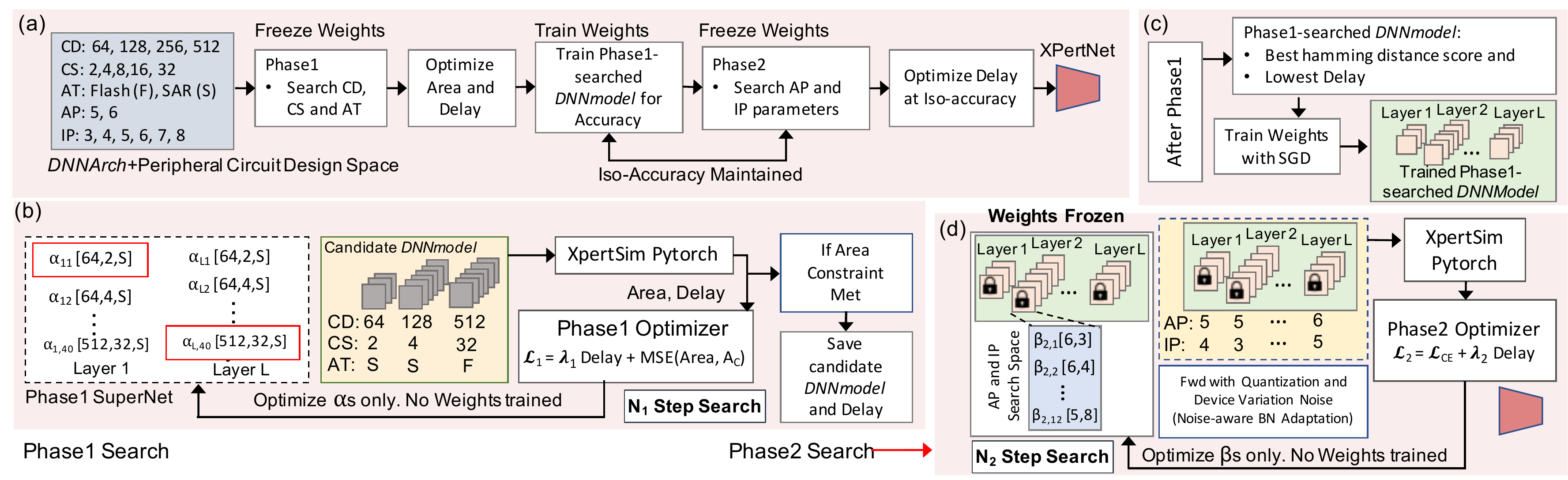} \\ 
    \end{tabular}
    
    \caption{ (a) Overview of XPert's dual-phase co-search (b) Phase1 co-search between CD, CS and AT to optimize area and delay. $A_C$- user provided area constraint. (c) Post-Phase1 training of the Phase1-searched \textit{DNNmodel} (d) Phase2 co-search to optimize hardware accuracy and to further reduce the delay. }
    \label{fig:phase1_phase2}
\end{figure*}

Fig. \ref{fig:phase1_phase2}a shows the \textit{DNNArch}-peripheral circuit design space and a summary of XPert's dual-phase co-search approach. XPert's \textit{DNNArch} design space consists of layer-specific channel depths only as it significantly contributes to the EDAP and accuracy of an IMC-implemented DNN (see Section \ref{sec:background_motivation}).

\textbf{Phase1 Co-search: }The Phase1 SuperNet consists of L layers, with each layer containing different combinations of CD, CS and AT values. In XPert, layer $l$ in the Phase1 SuperNet contains 4$\times$5$\times$2= 40 nodes with associated architectural parameters $\alpha_{l,1}$-$\alpha_{l,40}$. In each search step, node $i$ with the highest $softmax(\alpha_{l,i})$ value is sampled (shown with red boxes in Fig. \ref{fig:phase1_phase2}b) resulting in a candidate \textit{DNNmodel} with layer-specific CD, CS and AT configurations. Note, during Phase1, the AP and IP configurations are 6 and 8, respectively for all layers. The candidate \textit{DNNmodel} is sent as an input to the XPertSim-Pytorch tool. XPertSim-Pytorch contains GPU-compatible hardware-accurate differentiable area and delay functions for real-time loss $\mathcal{L}_1$ computation and gradient backpropagation. $\mathcal{L}_1$ is a combination of the delay and mean squared error loss (MSE) between the sampled \textit{DNNmodel's} area and the user provided area constraint. The gradients $\frac{\partial \mathcal{L}_1}{\partial \alpha_{l,i}}$ are used to optimize the $\alpha_{l,i}$ using standard SGD. Note, during Phase1, we only update the $\alpha$s while the weights are not trained. At each step, if the area of any candidate \textit{DNNmodel} is nearly equal (within 2\% margin) to the user provided area constraint, the candidate \textit{DNNmodel} and its delay are saved. The Phase1 search occurs over $N_1$ steps.

\textbf{Post Phase1 Co-search:} As shown in Fig. \ref{fig:phase1_phase2}c, among the pool of candidate \textit{DNNmodels} (saved during Phase1), we select the candidate with the highest Hamming Distance (HD) score \cite{mellor2021neural} and lowest delay. We call this candidate the Phase1-searched \textit{DNNmodel}. The HD score measure is adopted from a recent Neural Architecture Search work \cite{mellor2021neural} which uses HD as a fast and indirect metric to determine the classification capability of different subnetworks sampled from a supernetwork without the need to train the architecture/weight parameters\footnote{inputs are passed through an untrained DNN and the hamming distance between unique binary code representations of the ReLU maps is carried out. Higher hamming distance score represents higher representation capability.}. The weights of the Phase1-searched \textit{DNNmodel} are trained till convergence using standard SGD. 

\textbf{Phase2 Co-search:} The weights, CD, CS and AT configurations of the trained Phase1-searched \textit{DNNmodel} are frozen and only the layer-specific AP and IP are co-searched. For each layer of the Phase1-searched \textit{DNNmodel}, there are 2$\times$6=12 different combinations of AP and IP as shown in Fig. \ref{fig:phase1_phase2}d with architectural parameters $\beta_{l,i}$  for layer $l$ and node $i$. Like Phase1, the nodes with the highest $softmax(\beta_{l,i})$ are sampled. A forward propagation is performed on the trained Phase1-searched \textit{DNNmodel} with the selected layer-specific AP and IP configurations. Since, during Phase2, the AP and IP can be lower than 6 and 8, respectively (default values in Phase1), quantization noise is induced during forward propagation. Additionally, we also include device variation noise during the forward propagation. To mitigate the quantization/device variation noise, we use batchnorm adaptation \cite{tsai2020robust} wherein the running mean of the batch-normalization layers are updated based on the noisy convolution outputs. Through the forward propagation, the cross-entropy loss $\mathcal{L}_{CE}$ is computed. The batchnorm adaptation is performed over 20k randomly sampled images from the training set. The delay of the trained Phase1-searched \textit{DNNmodel} is computed by  XPertSim-Pytorch based on the selected AP/IP configurations. $\mathcal{L}_{CE}$ and delay are used to compute the loss function $\mathcal{L}_2$. Gradients $\frac{\partial \mathcal{L}_2}{\partial \beta_{l,i}}$  are used to optimize the $\beta_{l,i}$s using SGD. Phase2 occurs over $N_2$ steps.


\begin{figure}[t]
    \centering
         \includegraphics[width=1\linewidth]{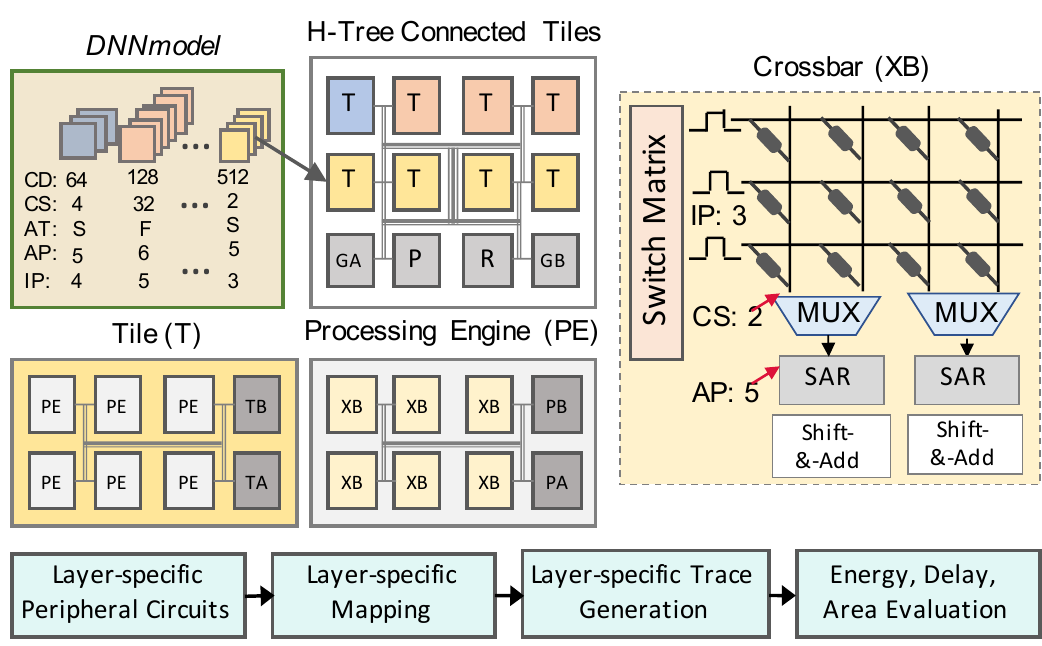}
    \caption{\textit{DNNmodel} mapping on the XPertSim C++ platform and tool flow. GB, TB, PB- Global, Tile and PE Buffers. GA, TA, PA- Global, Tile and PE Accumulators, P- Pooling and R-ReLU modules. Switch matrix provide voltage inputs to the crossbar. Shift-\&-Add, PA, TA and GA accumulate partial sum outputs. PB, TB and GB store intermediate MAC and activation values.}
    \label{fig:XPertSim}\vspace{-4mm}
\end{figure}

\textbf{XPertSim-C++} is a cycle-accurate hardware evaluation platform adopted from Neurosim \cite{peng2019dnn+} to support IMC-implemented DNNs with heterogeneous peripheral circuit configurations. XPertSim-C++ serves as a backend for XPertSim-Pytorch. Fig. \ref{fig:XPertSim} shows a \textit{DNNmodel} with layer-specific CD, CS, AT, AP and IP being mapped on a tiled architecture. The Tiles, PEs and crossbars use H-Tree-based communication. The crossbar's configuration depends on the circuit parameters of the mapped layer. Post mapping, XPertSim C++ performs hardware-accurate energy, delay, area evaluation. 



\section{Experiments and Results}
\subsection{Experimental Setup}\vspace{-3mm}
\begin{table}[h!]
    
    \centering
    \caption{IMC Hardware Implementation Parameters.}
    \label{tab:hw_parameter}
    \resizebox{\linewidth}{!}{
    \begin{tabular}{|c|c|} \hline
        
        
        Technology & 32nm CMOS  \\ \hline
        Device & 4-bit RRAM \cite{hajri2019rram} ($\sigma$/$\mu$ = 20\%) \\ \hline
        $R_{on}$ (on/off ratio) & 6 k$\Omega$ (150) \\ \hline
        Crossbar Size (X), \#Xbars/Tile & 64, 64 \\ \hline
        Weight Quantization, Weight \& Input Slicing & 8-bit, 4-bit \& 1-bit \\ \hline
    \end{tabular}}

\end{table}
We consider CIFAR10 and TinyImagenet datasets to evaluate the efficacy of the XPert platform.

\textbf{Design Space:} The \textit{DNNArch}+Peripheral circuit design space is shown in Fig. \ref{fig:phase1_phase2}a. XPert's \textit{DNNArch} has a VGG16 backbone with 13 convolution and 1 fully-connected (FC) layers with 3$\times$3 kernel. The CDs are co-searched with peripheral circuits. The \textit{DNNArch} of the baseline, is a standard VGG16 network\footnote{adopted from https://github.com/kuangliu/pytorch-cifar} with 13 convolution and 1 FC layer.

\textbf{Phase1 and Phase2 Parameters: } The Phase1 and Phase2 co-search is conducted for 2000 and 20 steps, respectively on Pytorch 1.1.0 with Nvidia RTX-2080ti GPU backend. Since the weights are not trained during co-search, the Phase1 and Phase2 require merely 5 and 10 GPU minutes which is negligible compared to the training time of the Phase1-searched \textit{DNNmodel}. $\lambda_1$ and $\lambda_2$ are 0.01 and 0.001, respectively. Learning rates for Phase1 and Phase2 SGD-based co-search are 13 and 0.1, respectively. Post Phase1 co-search, the Phase1-searched \textit{DNNmodel} is trained for 200 epochs on the respective dataset using SGD with a learning rate of 0.1 with cosine decay. Hardware evaluations of XPertNets and VGG16 baselines are performed on the XPertSim C++ platform with parameters shown in Table \ref{tab:hw_parameter}. Further, we perform quantization/device noise-aware batchnorm adaptation of VGG16 baselines to obtain competitive NI accuracy on IMC implementation. 



\subsection{Comparison with Homogeneous VGG16 Implementations}
\begin{table*}
\centering
\captionsetup{justification=centering}
\caption{Table comparing XPertNets with homogeneous VGG16 implementations for CIFAR10 and TinyImagenet datasets. The notations used are same as in Fig. \ref{fig:first_result_fig}. All implementations are on 64$\times$64 4-bit RRAM crossbars. For VGG16, AP/IP are 6 and 8, respectively. I- Ideal accuracy without any hardware noise. NI- Non-ideal accuracy with quantization and device variation noise. $XPertNet_{D, 50}$- XPertNet searched for dataset D at 50mm$^2$ area constraint.}
\label{tab:comp_VGG16}
\begin{tabular}{|c|c|c|c|c|c|c|c|c|} \hline
    \multirow{2}{*}{Configuration} & \multirow{2}{*}{Area (mm$^2$)} & \multirow{2}{*}{Delay (ns)} & \multirow{2}{*}{Energy (pJ)} & \multirow{2}{*}{\begin{tabular}{c}
         EDAP \\
         (mJ$\times$ms$\times$mm$^2$)
    \end{tabular}} & \multirow{2}{*}{TOPS/W} & \multirow{2}{*}{TOPS/mm$^2$} &
    \multicolumn{2}{|c|}{Accuracy} \\ \cline{8-9}
     & & & & & & & I & NI\\ \hline
    \multicolumn{9}{|c|}{\textbf{CIFAR10}} \\ \hline 
    VGG16 [F+CS16] & 99 & 3.75e+06 & 8.02e+07 & 29.72 & 21.54 & 0.0062 & 93.97 & 82.7\\ \hline

    VGG16 [F+CS8] & 135 & 3.19e+06 & 7.01e+07 & 30.14 & 24.18 & 0.005 & 93.97 & 82.7\\ \hline
 
    VGG16 [F+CS32] & 81 & 5.04e+06 & 1.08e+08 & 43.74 & 16.09 & 0.0053 & 93.97 & 82.7\\ \hline

    VGG16 [F+CS4] & 203 & 4.27e+06 & 6.98e+07 & 60.50 & 22.63 & 0.0025 & 93.97 & 82.7\\ \hline

    
    $XPertNet_{C, 50}$ & \textbf{50.2} & 2.16e+06 & \textbf{2.69e+07} & \textbf{2.9} & 35.7 & 0.01 & {93.6} & \textbf{92.46} \\ \hline
     $XPertNet_{C, 60}$ & 59.6 & \textbf{2.15e+06} & 3.32e+07 & 4.28 & \textbf{37.24} & \textbf{0.012} & \textbf{94.1} & 92.11 \\ \hline
     
     \multicolumn{9}{|c|}{\textbf{TinyImagenet}} \\ \hline 
     VGG16 [F+CS16] & 95 & 9.01e+06 & 1.48e+08 & 126.5 & 16.1 & 0.0035 & 60 & 52\\ \hline
    
    VGG16 [F+CS8] & 137 & 7.76e+06 & 1.32e+08 & 140.3 & 17.57 & 0.002 & 60 & 52\\ \hline
    
    VGG16 [F+CS32] & 82.4 & 1.19e+07 & 1.95e+08 & 191.2 & 12.38 & 0.003 & 60 & 52\\ \hline
    
    VGG16 [(F+CS4] & 205 & 1.03e+07 & 1.34e+08 & 282 & 15.91 & 0.0014 & 60 & 52\\ \hline
    
    $XPertNet_{T,50}$ & \textbf{49.5} & 5.91e+06 & \textbf{8.98e+07} & \textbf{26.53} & 26.08 & 0.0096 & {58} & \textbf{56.7} \\ \hline
     $XPertNet_{T,60}$ & 60.3 & \textbf{4.88e+06} & 9.28e+07 & {27.17} & \textbf{26.14} & \textbf{0.0107} & 59 & 56.3\\ \hline
\end{tabular}\vspace{-5mm}
\end{table*}

    
     
    
Table \ref{tab:comp_VGG16} shows the hardware efficiency and accuracy results of XPertNets and 4 homogeneous VGG16 implementations with the lowest EDAP values. For CIFAR10, $XPertNet_{C,50}$ (shown in Fig. \ref{fig:XPertNet_config}a) having an area of 50mm$^2$ achieves 2.98$\times$ and 10.24$\times$ lower energy and EDAP, respectively than VGG16 [F+CS16]. $XPertNet_{C,50}$ also achieves the best NI accuracy of 92.46\% (9.7\% higher than VGG16 [F+CS16]). $XPertNet_{C,60}$ (shown in Fig. \ref{fig:XPertNet_config}a) achieves 1.72$\times$ and 1.93$\times$ higher TOPS/W and TOPS/mm$^2$ than VGG16 [F+CS16]. For TinyImagenet, $XPertNet_{T,50}$ achieves 4.7\% higher NI accuracy at 4.7$\times$ lower EDAP than VGG16[F+CS16]. $XPertNet_{T,60}$ achieves 1.62$\times$ and 3$\times$ higher TOPS/W and TOPS/mm$^2$ than VGG16 [F+CS16].

\begin{figure}[t]
    \centering
         \includegraphics[width=0.9
         \linewidth]{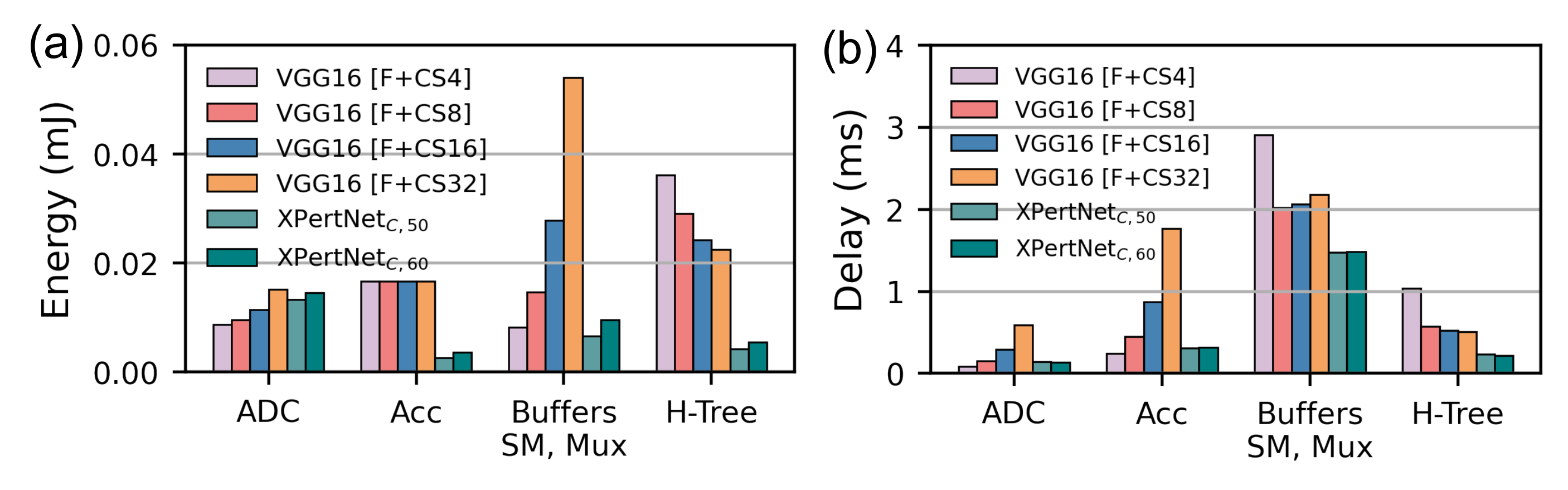}
    \caption{Plots showing the (a) CIFAR10 energy (b) CIFAR10 delay of different IMC components for VGG16,  $XPertNet_{C, 50}$ and $XPertNet_{C, 60}$ \textit{DNNmodels}. Acc- Accumulators and SM- Switch Matrices and Mux- Multiplexers. }
    \label{fig:ED_Distribution}\vspace{-4mm}
\end{figure}

\textbf{Energy Consumption Analysis} From Fig. \ref{fig:ED_Distribution}a and Fig. \ref{fig:ED_Distribution}b, we find that the $XPertNet_{C,50}$ and $XPertNet_{C,60}$ entail significantly low  ADC, Accumulation (Acc), data communication (H-Tree), storage (Buffers) and switch matrix energy/delay which results in a high energy efficiency. Similar observations are made in case of TinyImagenet.

\subsection{Comparison with Prior Works}\vspace{-2mm}

\begin{table}[h!]
    \centering
    \captionsetup{justification=centering}
    \caption{Performance comparison of prior IMC-aware \textit{DNNArch} co-search works and XPert (results correspond to $XPertNet_{C,50}$). S-Searched Parameter, NS- Not Searched, $X$- crossbar size}
    \label{tab:prior_NAS_comp}
    \resizebox{0.9\linewidth}{!}{
    \begin{tabular}{|c|c|c|c|} \hline
         Work & Jiang \textit{et al.} \cite{jiang2020device} & NAX \cite{negi2022nax} & \textbf{XPert (Ours)} \\ \hline
         Backbone Network & VGG11 & ResNet20 & VGG16 \\ \hline
         Channel Depth & S & NS & \textbf{S} \\ \hline
         Kernel Size & S & S & \textbf{NS (3$\times$3)}\\ \hline
         Crossbar Size & NS (64$\times$64) & S & \textbf{NS (64$\times$64)}\\ \hline
         Weight Quantization & S & NS (8-b) & \textbf{NS (8-b)} \\ \hline
         Layer Input Precision & S & NS (8-b) & \textbf{S} \\ \hline
         ADC Precision & NS (4-b) & NS $log_2(3X)$ & \textbf{S} \\ \hline
         ADC Type & NS & NS & \textbf{S} \\ \hline
         Column Sharing & NS & NS & \textbf{S} \\ \hline
         Device & 4-b RRAM & 2-b RRAM & \textbf{4-b RRAM} \\ \hline
         Accuracy (CIFAR10) & 93.12 & 92.7 & \textbf{92.46}  \\ \hline
         EDAP (mJ$\times$ms$\times$mm$^2$) & 2631 & 290 & \textbf{2.9} \\ \hline
         \multirow{2}{*}{$\frac{Baseline ~EDAP}{Optimized ~EDAP}$}& \multirow{2}{*}{0.24} & \multirow{2}{*}{1.2} & \multirow{2}{*}{\textbf{10.24}} \\
         
         & & & \\\hline
    \end{tabular}}
    
\end{table}\vspace{-2mm}


Table \ref{tab:prior_NAS_comp} compares different IMC-aware \textit{DNNArch} co-search works with XPert at iso-accuracy for CIFAR10 dataset. As different works use different baselines and hardware evaluation platforms, we use the $\frac{Baseline ~EDAP}{Optimized ~EDAP}$ ratio to perform a fair hardware-efficiency evaluation. In \cite{jiang2020device}, the authors co-searched the channel depths, kernel size and layer-specific quantization of a VGG11 backbone network to achieve optimal accuracy. Although they achieved 3\% higher CIFAR10 accuracy compared to the VGG11 baseline, the EDAP of the optimized network was 4.16$\times$ higher than the baseline (causing $\frac{Baseline ~EDAP}{Optimized ~EDAP} <$ 1). In NAX \cite{negi2022nax}, the authors performed kernel and IMC crossbar size co-search to achieve high accuracy (92.7) with 1.2$\times$ lower EDAP than a ResNet20 baseline. Xpert ($XPertNet_{C,50}$), achieves a $\frac{Baseline ~EDAP}{Optimized ~EDAP}$ of 10.24$\times$ at 92.46\% accuracy attributed to the co-search of \textit{DNNArch} and peripheral circuit parameters. \vspace{-1.5mm}




\subsection{Analysing XPertNets at Different Area Constraints}
\label{sec:XPertNets}
\begin{figure}[t]
    \centering
    \includegraphics[width=\linewidth]{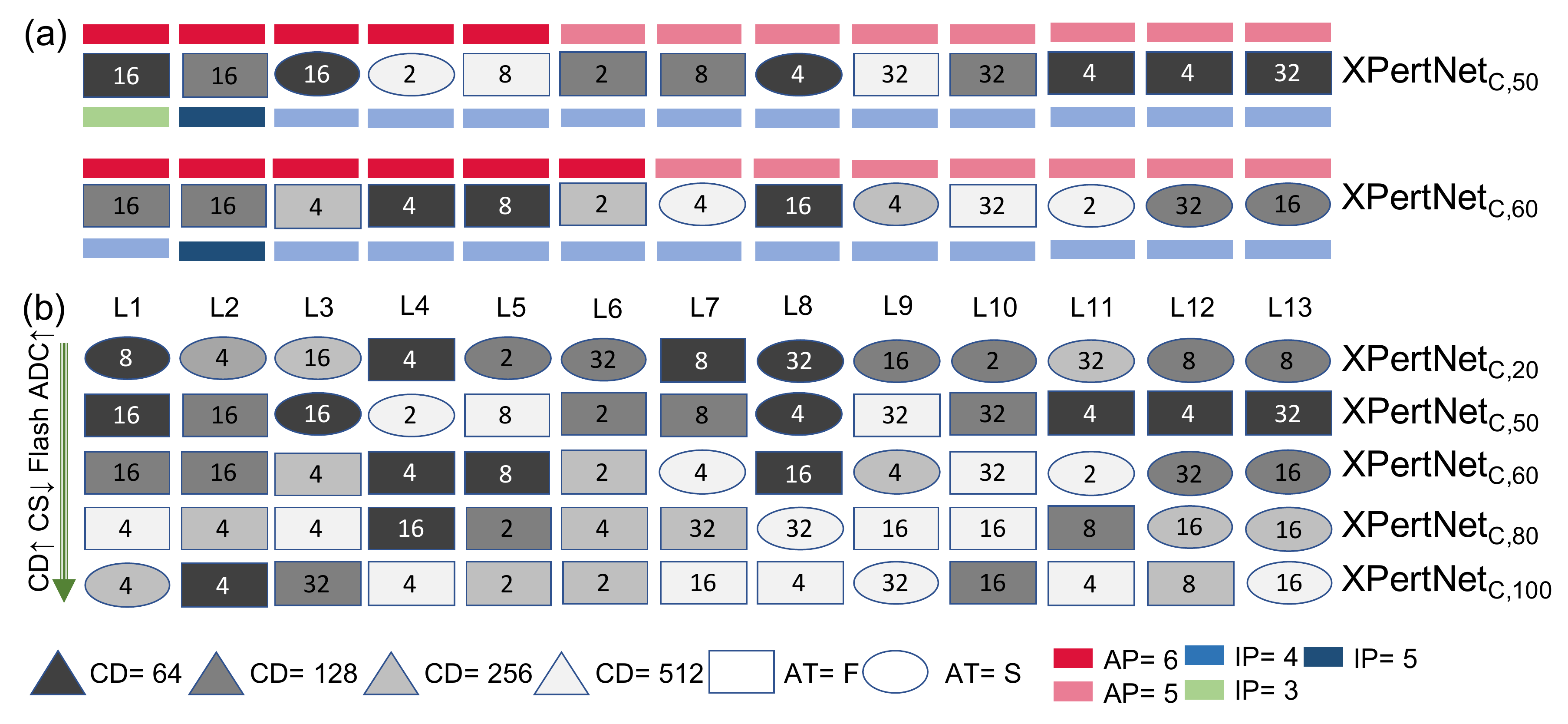}
    \caption{(a) $XPertNet_{C,50}$ and $XPertNet_{C,60}$ configurations. (b) XPertNet configurations (only convolution layers) under different area constraints for CIFAR10 dataset. Numbers inside the shapes denote CS values.}
    \label{fig:XPertNet_config}\vspace{-4mm}
\end{figure}\vspace{-2mm}
\begin{figure*}[t]
    \centering
    \includegraphics[width=0.7\textwidth]{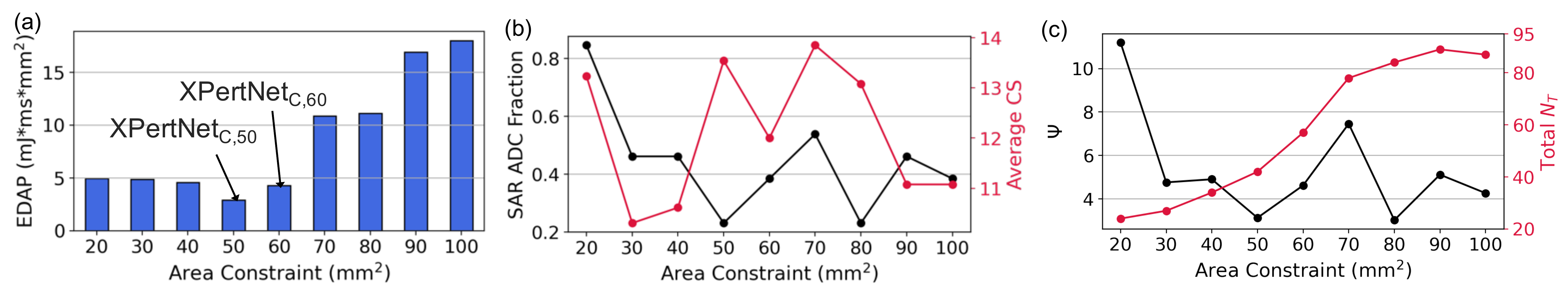}
    \caption{(a) EDAP trend for CIFAR10-based XPertNets at iso-accuracy (b) SAR ADC fraction and average CS in XPertNets and (c) $\Psi$ and total $N_T$ of XPertNets obtained at different area constraints. Total $N_T$ is the total number of tiles consumed by a particular XPertNet.}
    \label{fig:ED_performance_search}\vspace{-6mm}
\end{figure*}
\begin{figure}[t]
    \centering
    \includegraphics[width=0.8\linewidth]{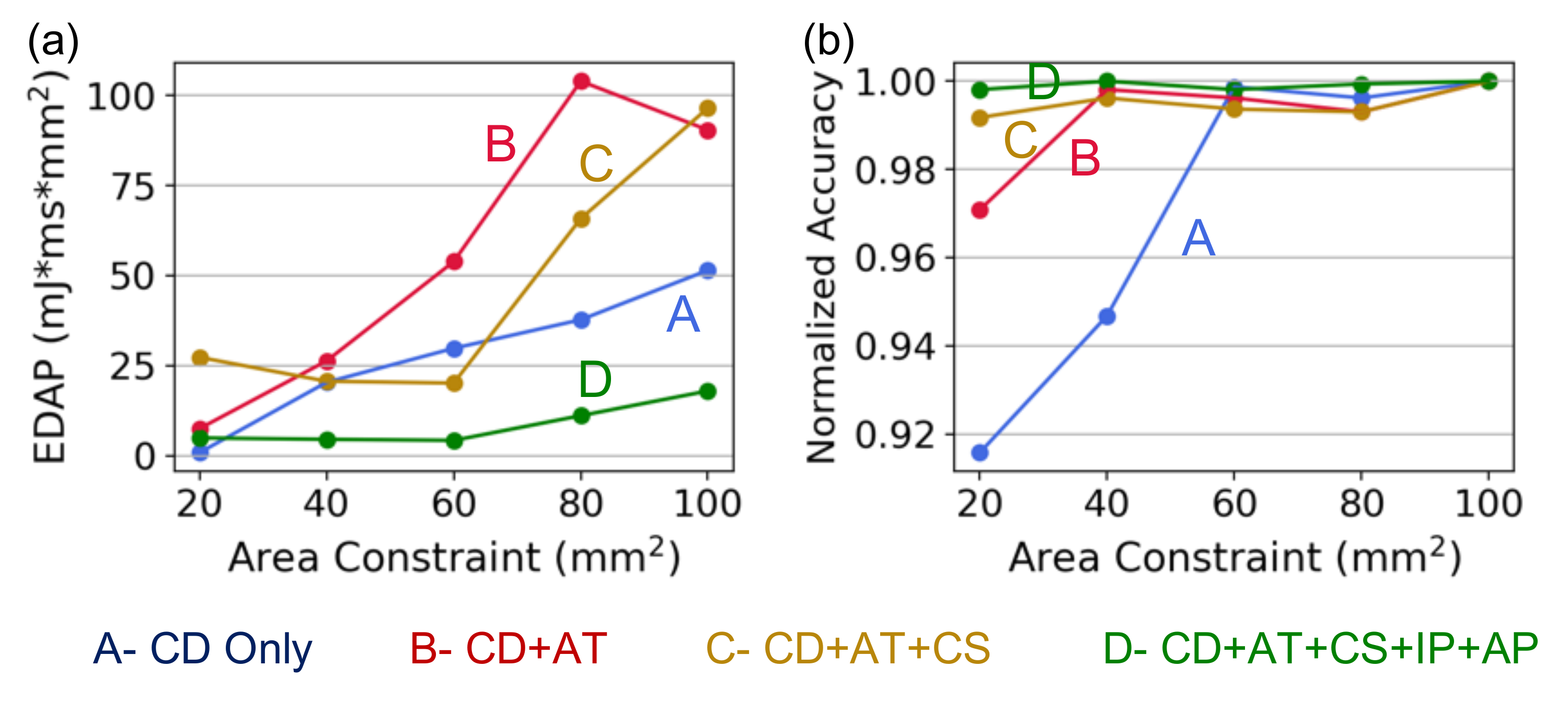}\vspace{-1mm}
    \caption{Figure showing the area-wise (a) EDAP and (b) normalized accuracy (with respect to the highest achieved accuracy) when different \textit{DNNArch} and peripheral circuit parameters are co-searched.}
    \label{fig:design_space_ablation}
    \vspace{-4mm}
\end{figure}
\begin{figure}[t]
    \centering
    \includegraphics[width=0.7\linewidth]{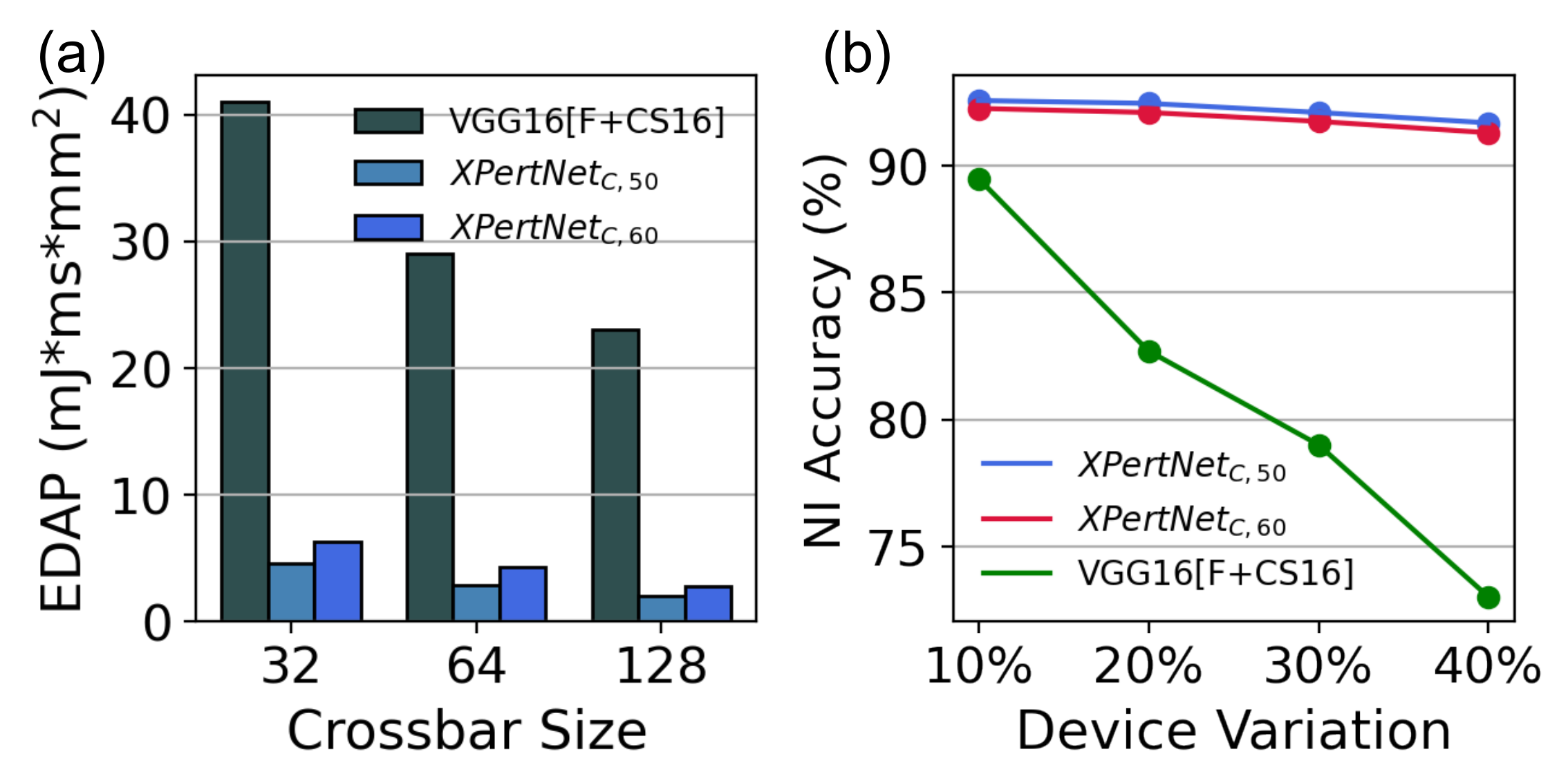}
    \caption{Comparison of (a) EDAP for different crossbar sizes (b) NI accuracy under different device variations between XPertNet and VGG16[F+CS16].}
    \label{fig:xbar_and_d_variations}\vspace{-2mm}
\end{figure}
Interestingly, as seen in Fig. \ref{fig:XPertNet_config}a, to achieve optimal accuracy, XPert assigns higher AP and IP in the shallow layers (for better representation of input features) while reduces the AP/IP configurations in the deeper layers to increase energy-efficiency. The AP/IP trends are similar across other XPertNets and hence not shown. Fig. \ref{fig:XPertNet_config}b shows the CIFAR10-based XPertNet configurations obtained under different area constraints. As the area constraint increases, the CD and the number of Flash ADCs in the XPertNets increase (reduction in the number of circles and more lighter shades) while the CS decreases. Similar trends are observed for XPertNets searched for TinyImagenet dataset. 

\textbf{EDAP at Different Area Constraints:}
Interestingly, as seen in Fig. \ref{fig:ED_performance_search}a, the minimum EDAP is obtained at area constraints of 50mm$^2$ and 60mm$^2$. To explain this, we define $\Psi$ (plotted in Fig. \ref{fig:ED_performance_search}c) as the product of average CS and SAR ADC fraction (plotted in Fig. \ref{fig:ED_performance_search}b). $\Psi$ empirically determines the delay incurred due to the crossbar read. Higher number of SAR ADCs and average CS in an XPertNet increases the number of clock cycles for crossbar read and hence more delay (also shown in Fig. \ref{fig:latency_vs_params}b). At area $\leq$40mm$^2$, $N_T$ is small but the $\Psi$ value is high due to high average CS and SAR ADC fraction. This results in higher EDAP. For area $\geq$ 70mm$^2$, although $\Psi$ decreases, the delay increases as XPertNets have larger CDs (as seen in Fig. \ref{fig:XPertNet_config}b) resulting in higher Total $N_T$. \vspace{-1mm}

\subsection{Significance of Different Design Spaces}\vspace{-1mm}
Fig. \ref{fig:design_space_ablation} shows the significance of co-search across different \textit{DNNArch} and peripheral circuit design spaces. 
In \textbf{A}, where only CDs are searched with fixed peripheral circuit configurations (CS=8, AT=F, AP=6 and IP=8), the \textit{DNNArchs} sampled at higher area constraint have high EDAP and accuracy due to large \textit{DNNArch} size that accommodates more weight parameters. 
At low area constraints, the EDAP is low at the cost of low accuracy. In \textbf{B}, when CD is co-searched with AT (at CS=8, AP=6, IP=8), the accuracy at low area constraints increases as more weight parameters can be accommodated with the help of area-efficient SAR ADCs. In \textbf{C}, when CD, AT and CS are co-searched (AP=6, IP=8),
the accuracy and EDAP at low area constraints further increases because SAR ADCs are now combined with high CS values which helps accommodate more weight parameters while also increasing the EDAP. Interestingly, at higher areas, the EDAP is lowered compared to design space B as smaller CS values are selected to achieve low delay and energy. Finally in \textbf{D}, when CD, AT, CS, AP and IP are holistically co-searched, the EDAP lowers significantly at high accuracy across different area constraints.
\subsection{Impact of Crossbar Sizes and Device Variation}\vspace{-1mm}

Fig. \ref{fig:xbar_and_d_variations}a, shows that $XPertNet_{C,50}$ and $XPertNet_{C,60}$ achieve significantly low EDAP compared to VGG16[F+CS16] irrespective of the crossbar size used for implementation. Additionally, XPertNets entail higher robustness against memristive device variations compared to VGG16[F+CS16] model as seen in Fig. \ref{fig:xbar_and_d_variations}b. This shows an interesting co-dependence between device variations and XPert-searched \textit{DNNmodel} configuration. \vspace{-1mm}

\section{Conclusion}\vspace{-1mm}
In this work, we design XPert, a framework that performs quick and optimized co-search in a large \textit{DNNArch}-peripheral circuit design space to obtain high accuracy, and low EDAP \textit{DNNmodels}. We show that a holistic co-search between CD, CS, AT, AP and IP is essential to attain high accuracy at low EDAP values. For CIFAR10 and TinyImagenet datasets, XPert achieves 10.24$\times$ and 4.7$\times$ lower EDAP, respectively while improving the NI-accuracy by 9.7\% and 4.7\%, respectively compared to homogeneous VGG16 implementations. \vspace{-2mm}

\small
\section*{Acknowledgement}\vspace{-1mm}
This work was supported in part by CoCoSys, a JUMP2.0 center sponsored by DARPA and SRC, Google Research Scholar Award, the National Science Foundation CAREER Award, TII (Abu Dhabi), the DARPA AI Exploration (AIE) program, and the DoE MMICC center SEA-CROGS (Award \#DE-SC0023198).
\normalsize





\bibliographystyle{IEEEtran}

\bibliography{IMC_Co-search.bib}

\end{document}